\documentclass[10pt, a4paper]{article}
\usepackage{lrec-coling2024} 
\pdfoutput=1 

\usepackage{todonotes}
\usepackage{booktabs}
\title{Advancing Generative AI for Portuguese with Open Decoder\\Gervásio~PT*}

\name{Rodrigo Santos, João Silva, Luís Gomes, João Rodrigues, António Branco} 

\address{
    University of Lisbon
    \\
        NLX - Natural Language and Speech Group, Department of Informatics\\
    Faculdade de Ciências,
    Campo Grande, 1749-016 Lisboa, Portugal\\
    \{rsdsantos, jrsilva, luis.gomes, jarodrigues, antonio.branco\}@fc.ul.pt\\}

\abstract{ % abstract of 150 to 200 words
To advance the neural decoding of Portuguese, in this paper we present a fully open Transformer-based, instruction-tuned decoder model that sets a new state of the art in this respect.
To develop this decoder, which we named Gervásio~PT*, a strong LLaMA~2 7B model was used as a starting point, and its further improvement through additional training was done over language resources that include new instruction data sets of Portuguese prepared for this purpose, which are also contributed in this paper.
All versions of Gervásio are open source and distributed for free under an open license, including for either research or commercial usage, and can be run on consumer-grade hardware, thus seeking to contribute to the advancement of research and innovation in language technology for Portuguese.
 \\ \newline  }

\begin{document}

\maketitleabstract

%\todo[inline]{Up to eight (8) pages, excluding any number of additional pages for references, ethical considerations, conflicts-of-interest, as well as data, and code availability statements. For the final versions, authors of accepted papers will be given 1 extra content page to take the reviews into account.}
\section{Introduction}

This paper presents a model that is the first competitive, 7~billion parameter, fully open and fully documented large language model of the decoder family of Transformers that is prepared specifically for the Portuguese language, by means of instruction tuning, for both the European variant, spoken in Portugal (PTPT) and the American variant, spoken in Brazil (PTBR). By being fully open, it is open source and openly distributed for free under a free license, including for research and commercial purposes. By being fully documented, the new datasets that were specifically developed for its construction can be reused, its development can be reproduced, and reported performance scores can be independently assessed. By being fully open and documented, its further development and improvement is openly available to the community.

In the last half decade, the neural approach to natural language processing became pervasive, with virtually any language processing task attaining top performance under the Transformer architecture \citep{Vaswani:2017:Transformer}.
Initially proposed and explored in an encoder-decoder setup \citep{Raffel:2020:T5}, subsequent research has shown the particular strengths of separate encoder-only and decoder-only solutions \citep{Devlin:2019:BERT,He:2021:DeBERTa,Brown:2020l:GPT3}, with decoders becoming specially notable with the availability of ChatGPT to the general public \citep{Ouyang:2022:InstructGPT,openai2023gpt4}.

Among the thousands of natural languages spoken in the world, English is the one whose research is, by a huge margin, better funded and thus the one for which more language resources exist, including the gigantic collections of text that are necessary to train top performing large language models.
Consequently, the largest and best performing monolingual models have been developed for this particular language \citep{Touvron:2023:Llama2,He:2021:DeBERTa}.

Seeking to build on the strength of such monolingual models, multilingual models have also been developed. 
Typically, they are trained over datasets where relatively small portions of data from a few other languages are added to the data from English \citep{Devlin:2019:BERT,Scao:2022:Bloom}.
Interestingly, these models have shown competitive performance in handling tasks in languages other than English, leveraged by the massive volume of data thus made available and outdoing the meager results that would otherwise be obtained if a monolingual model had been trained only in the data available for those languages alone \citep{pires2019multilingual}.

In order to further mitigate the relative data scarceness impacting the non-English languages, further approaches have been undertaken that include the continuation of the self-supervised training with monolingual data from a specific language.
This continuation of causal language modelling (CLM) has been experimented with over multilingual models or even monolingual English models.
Research has shown that when such training is appropriately continued, the performance of the resulting model for that specific language exceeds the performance of the baseline model on that language, whose training has not been thus continued \citep{kaplan2020scaling,Rodrigues:2023:Albertina}.

By exploiting this approach of continuing the training of a previous strong foundation model, we contribute a new model with instruction tuning to foster the technological preparation of the Portuguese language.
To the best of our knowledge, this is the first decoder under the Transformer architecture that is both
(i)~specifically improved for Portuguese, covering two variants of this language, namely PTBR and PTPT,
%namely American Portuguese, spoken in Brazil, and European Portuguese, spoken in Portugal,
and (ii)~fully open, that is it cumulatively complies with all the features of being open source and openly distributed for free under a most permissive license (including for research and for commercial purposes).
The model is available at \url{https://huggingface.co/PORTULAN}.

%With 7~billion parameters, to the best of our knowledge and at the time of the submission of this paper, it represents the state of the art reported in the literature for the neural decoding of Portuguese, as it is the best performing decoder in mainstream benchmarks for this language, with evaluation scores coming close\todo{JS: Confirmar se assim se mantém} to performance scores in similar tasks reported for much larger encoders for Portuguese, with 65~billion parameters\footnote{Scores reported in pre-print \citep{pires2023sabia}, but not peer-reviewed neither publicly distributed for reuse or scientific reproduction.}

 To the best of our knowledge and at the time of writing, Gervásio represents the state of the art reported in the literature for open, 7~billion parameter decoders for Portuguese, surpassing the model it is based on as well as other decoders for Portuguese of similar size.
The release of Gervásio, alongside the instruction dataset used to train it and which is also a novel contribution of this paper, seeks to contribute to foster research and innovation for the language technology for Portuguese.

The remainder of this paper is organized as follows:
Related work is covered in the next Section~\ref{sec:rel_work}; 
the data used to train and test the model is presented in Section~\ref{sec:data};
Section~\ref{sec:models} describes the decoder for Portuguese created in this study and
Section~\ref{sec:results} presents and discusses the results of its evaluation.
The last Section~\ref{sec:conclusion} offers concluding remarks.

\section{Related work}
\label{sec:rel_work}

In this section we discuss previous results and resources in the literature that are related to the aim of the present paper.
We first address decoders for Portuguese that are publicly reported or publicly distributed, and then we address the available options concerning the base model that can be used to be continued to be trained with Portuguese data.

\subsection{Decoders for Portuguese}
\label{sec:enc_pt}

%\todo[inline]{Secção desatualizada, o que afeta a narrativa do paper. No HuggingFace temos, por exemplo, Boana (LLaMa2, 7B), Bode (LLaMa2, 7B e 13B), Cabrita (OpenLLaMA, 3B), Cabra (Mistral, 7B e 13B), Canarim (LLaMa2, 7B) e Samba (TinyLlama, 1.1B). Existe ainda o Glória, publicado a 20/02 no arXiv.}

%\todo[inline]{Relevantes: Sabiá, que já está publicado e disponibilizado; e o Bode}

Looking for decoders specifically developed or improved for Portuguese that are publicly distributed and for which it is possible to find a publicly available report, to the best of our knowledge there can be found only two that, with 7~billion parameters or more, match or surpass the size of Gervásio~PT* contributed in the present paper, namely the Sabiá models with 7 and 65 billion parameters \citep{pires2023sabia}.
%and the Bode \citep{garcia2024introducing} models. 
It is worth noting that:
(i)~these two models were developed for only one of the variants of Portuguese, PTBR, but not for PTPT;
(ii)~the 65 billion parameter model is reported in that publication but it is not distributed;
and (iii)~the 7~billion parameter model is distributed in a non open license, being its reuse restricted to research purposes only.

Other decoders that at the time of writing the present paper can be found of comparable size are not documented, besides being also for only one of the variants of Portuguese, namely PTBR:
Boana, Cabra, Cabrita, Canarim.\footnote{All on HuggingFace, at lrds-code/boana-7b-instruct, nicolasdec/CabraMistral7b-0.2, 22h/open-cabrita3b, and dominguesm/canarim-7b, respectively.}

% Boana-7B-Instruct: \url{https://huggingface.co/lrds-code/boana-7b-instruct}
% Cabra: \url{https://huggingface.co/nicolasdec/CabraMistral7b-0.2}
% Cabrita: \url{https://huggingface.co/22h/open-cabrita3b}
% Canarim: \url{https://huggingface.co/dominguesm/canarim-7b}

The other decoders, numbering about a dozen, that can be found for Portuguese have a smaller size, and are also only for PTBR.
The largest of these, the 3~billion parameter Cabrita mentioned above, is distributed through Hugging Face (HF) and documented in a non peer-reviewed publication \citep{larcher2023cabrita}.
%The largest of these, with 3~billion parameters and named Cabrita~3B, is distributed in Hugging Face (HF) and documented in a non peer-reviewed publication \citep{larcher2023cabrita}.
The second largest is Aira,\footnote{On HF at nicholasKluge/Aira-2-portuguese-1B7.} with 1.7~billion parameters and based on Bloom.
%The second largest is Aira\footnote{\url{https://huggingface.co/nicholasKluge/Aira-2-portuguese-1B7}} and it is presented in HF as having 1.7~billion parameters and being based on Bloom.
No evaluation results on benchmarks or downstream tasks for it are reported, it has a residual number of downloads from HF and, being based on Bloom, it inherits the restrictions from Bloom's license and it is thus not fully open as Gervásio.

%The most popular of these models smaller than Gervásio, in turn, with over 900 downloads from HF last month, is named GPorTuguese-2.\footnote{\url{https://huggingface.co/pierreguillou/gpt2-small-portuguese}}
%It has 124~million parameters, but like Aira, no evaluation results in benchmarks or downstream tasks are reported for it.\todo{check}

Common to these decoders other than Sabiá, which are of similar or smaller size, is that while they are publicly distributed, no public detailed presentation of them seems to be provided, be it an implementation report or a paper, either in pre-print or in peer-reviewed versions.
This hampers knowing, among other aspects, which datasets were used for their training and thus hampers sensible comparison with other related work and models, which may risk being evaluated in datasets where they were trained.

Turning to Sabiá, while there is a paper with its reporting \citep{pires2023sabia}, this model was developed by a commercial company and the variant with 7~billion parameters is not openly distributed, with its license restricting its use only for research, a restriction inherited from the license of LLaMA~1 \citep{Touvron:2023:Llama}, which was taken as its base model.
The variant with 65 billion parameter, in turn, does not appear to be publicly distributed.
Sabiá is reported to have been obtained by continuing the training of LLaMA~1 both in its 7~billion and 65~billion parameter versions.
A third version of Sabiá was trained over GPT-J \citep{mesh-transformer-jax}, with 6~billion parameters.
All of these were trained for the PTBR variant of Portuguese only.

%Given the public unavailability of Sabiá, the performance scores reported for it cannot be reproduced by independent parties, and any comparison with other models of interest, such as Gervásio, has to be done only on the basis of the figures reported by the respective authors in its non peer-reviewed, pre-print.

Looking into the collection of tasks reported to have been used to evaluate Sabiá, one finds a few that are common with the evaluation of Gervásio, 
%namely BoolQ and BLUEX, from SuperGLUE, and SST2, from GLUE,
such as BoolQ, 
which were also machine translated into PTBR to evaluate Sabiá.
Additionally, Sabiá's authors present its performance scores in a few other downstream tasks whose datasets did not result from machine translation from English ones, but were developed originally in PTBR.

The performance scores from Sabiá's publication are repeated in Section~\ref{sec:results}, side by side with related scores of the Gervásio~PTBR, for American Portuguese.
Against this background, and as it will be discussed at length in that Section, at the time of this writing and to the best of our knowledge, Gervásio offers the state of the art in terms of \textbf{fully open} decoders specifically improved for Portuguese in both PTPT and PTBR variants, and it is the first 7~billion parameter decoder specifically developed and distributed for the PTPT variant.

\subsection{Base models}
\label{sec:base_models}

In this connection, it is worth noting also that not only Gervásio happens to be the top performing 7~billion open decoder for Portuguese, but also that it adopted the best possible setup and codebase available at the time of its development given the goals and requirements assumed for its construction.

There are a number of multilingual decoders reported in the literature, such as mBART \citep{liu-etal-2020-multilingual-denoising}, XLM-R \citep{conneau-etal-2020-unsupervised}, mT5 \citep{xue2021mt5}, ByT5 \citep{xue-etal-2022-byt5},  XGLM \citep{lin-etal-2022-shot}, mGPT \citep{shliazhko2023mgpt}, Bloom \citep{Scao:2022:Bloom}, and LLaMA \citep{Touvron:2023:Llama2}, to which the promising English open models Mistral \citep{jiang2023mistral} and Pythia \citep{biderman2023pythia} were added in our considerations of the options available.
From these possibilities, many had to be excluded given their non-open license, leaving only those from the Mistral, Bloom, Pythia, and LLaMA families as viable bases on which to build Gervásio.

From these, we decided to leave out Mistral given that, unlike the others, it is indicated to have been developed with no guardrails or other possible state-of-the-art preventive measures available that could help mitigate possible ethical issues.

From the remainder three models left, Bloom is distributed under a RAIL license,\footnote{\url{https://huggingface.co/spaces/bigscience/license}} which hampers its use in some important application domains, such as law and healthcare, and thus it was left aside.

Finally, as LLaMA models appear to generally deliver better performance than similarly sized Pythia models in the Hugging Face's Open LLM Leaderboard,\footnote{\url{https://huggingface.co/spaces/HuggingFaceH4/open_llm_leaderboard}} we adopted LLaMA for our base model. 
In this leaderboard, LLaMA appears as superior to all the other models mentioned above, except possibly to Mistral, for which it is a matching or close alternative option, with the important advantage over Mistral though of safeguarding ethical aspects to the extent possible given the current status of knowledge concerning foundation models.

\section{Data}
\label{sec:data}

In this section we present the datasets we developed or reused to train and evaluate Gervásio. 

\subsection{Developed datasets}
\label{sec:datadeveloped}

To benefit from the advantages of instruction tuning over standard supervised fine-tuning \citep{wei2022finetuned}, and to keep some alignment with mainstream benchmarks for English, we resorted to tasks and respective datasets in the GLUE \citep{Wang:2018:GLUE} and the SuperGLUE \citep{Wang:2019:SuperGLUE} collections.

\paragraph{Task selection} 

We selected those datasets where the outcome of their machine translation into Portuguese could preserve, in the target language, the linguistic properties at stake and thus be acceptable for the purposes of this paper. 

For instance, the COLA dataset from the GLUE benchmark contains examples of grammatical and non-grammatical expressions from English.
This dataset had to be put aside given that an automatic machine translator typically delivers grammatical expressions in the target language, even if the source expression is not grammatical, defeating the purpose of the benchmark.%\footnote{More details on the reasons to leave aside other datasets can be found in the Appendix of the accepted paper.}

From GLUE, we resorted to the following four tasks:
(i)~MRPC (paraphrase detection),
(ii)~RTE (recognizing textual entailment),
(iii)~STS-B (semantic textual similarity),
and (iv)~WNLI (coreference and natural language inference).
And from SuperGLUE, we included these other four tasks: (i)~BoolQ (yes/no question answering), (ii)~CB (inference with 3 labels),
(iii)~COPA (reasoning), and (iv)~MultiRC (question answering).

\paragraph{Task translation}
To machine translate into European Portuguese and into American Portuguese, we resorted to DeepL,\footnote{\url{https://www.deepl.com}} which was set for each one of these variants, respectively. %and into American Portuguese, to Google Translate.\footnote{\url{https://translate.google.com}}

\paragraph{Task templates}
Instruction templates have been manually crafted for each task.
These take the various fields in the dataset and arrange them into a prompt by, for instance, appending ``Frase 1:'' (Eng.~``Sentence 1:'') before the first sentence of an example in the RTE dataset.
A more detailed example is provided below in the Annex~A.
%These templates are listed in full detail in a dedicated webpage,\footnote{\url{https://huggingface.co/datasets/PORTULAN/extraglue}} of which one example is provided below in the Annex~A.

\begin{table}
    \centering
    \begin{tabular}{lrrr}
    \toprule
        task           & \#exs.tra  & \#exs.aug   & total   \\ \midrule
        STS-B          & 5749          & 5749        & 11498   \\
        WNLI           & 635           & 1270        & 1905    \\
        BoolQ          & 9427          & 28281       & 37708   \\
        CB             & 250           & 500         & 750     \\
        MultiRC        & 27243         & 81729       & 108972  \\ \midrule
        Total \#exs    & 43304         & 117529      & 160833  \\
        \midrule\addlinespace[1ex]\midrule
        %Total tokens   & 17947213      & 50133679    & 68080892\\
        Total \#tok pt & 17.9M      & 50.1M    & 68.0M\\
        Total \#tok br & 17.8M      & 50.6M    & 68.4M\\
    \bottomrule
    \end{tabular}
    \caption{Size of translated (tra) and augmented (aug) training datasets, in number of examples (\#exs). The number of examples is identical for both variants. Token counts (\#tok) concern examples only and do not include the instruction or the context examples in few-shot mode.}
    \label{tab:traindatastatistics}
\end{table}

\paragraph{Training data} 

For continuing causal language modelling (CLM) with Portuguese data, we used the datasets STS-B and WNLI, from GLUE, and BoolQ, CB and MultiRC, from SuperGLUE, machine translated into Portuguese twice, once for PTPT, and another time for PTBR.

For CLM, each training instance includes the task instruction followed by one or more examples taken from the training partition of that task (including the respective gold answers).

Every instance from the training partitions is seen twice during CLM:
once where it is the only example in the respective training instance (that is, it is not preceded by other examples --- zero-shot mode);
and once where it is preceded by other, 1 to \textit{n} randomly selected examples (few-shot mode), where \textit{n} is the largest number possible given the sequence length in CLM.\footnote{Exceptions were BoolQ and MultiRC, which given the size of their examples and the maximum sequence length of the model, allowed zero-shot mode only.}
Instances, examples, modes and values for \textit{n} are shuffled.

Statistics on the training datasets are in Table~\ref{tab:traindatastatistics}. %\todo{faltam algumas contagens}
Taking into account the instructions, the examples in few-shot mode and the two subsets, one for zero-shot mode and the other for few-shot mode, altogether, the CLM resorted to a 83~million token dataset (83.1M for PTPT and 83.6 for PTBR) when we trained our model.

% Corpus final PT-BR 83,053,220 tokens, 165559 examples?
% Corpus final PT-PT 83,622,122 tokens, 165559 examples?

\paragraph{Testing data} 

For testing, we reserved the translated datasets MRPC (similarity) and RTE (inference), from GLUE, and COPA (reasoning/qa), from SuperGLUE, which were taken as representatives of three major types of tasks, and were not seen during training in CLM.

Each testing prompt includes the task instruction followed by an instance from the validation partition (without the gold label).
This instance may be preceded by zero (in zero-shot prompting) or by a few examples (in few-shot prompting) taken from the training partition (these examples include the respective gold labels).

\paragraph{Augmented datasets}

Following \citep{iyer2023optiml}, we employ data augmentation techniques to enhance the size and diversity of our dataset.
This involves repurposing the tasks in various ways, such as generation of answers from MultiRC, question generation from BoolQ, and other relevant modifications.
These are presented in the Annex B.
Table~\ref{tab:traindatastatistics} summarizes the number of examples in the augmented datasets we arrived at.
We did not perform data augmentation for any dataset reserved for testing. 

\subsection{Reused datasets}

For further testing our decoder, in addition to the testing data described above, we also reused some of the datasets that had been resorted to by \citep{pires2023sabia} for American Portuguese to test the Sabiá model and that were originally developed with materials from Portuguese: ASSIN~2 RTE (entailment) and ASSIN~2 STS (similarity) \citep{assin2020realetal}, BLUEX (question answering) \citep{almeida2023bluex}, ENEM~2022 (question answering) \citep{nunes2023evaluating} and FaQuAD (extractive question-answering) \citep{faquad2019}.
To secure comparability with that model, we filtered out these datasets and prepared their test instances as indicated in the Annex of the Sabiá paper.\footnote{We did not reuse TweetSentBR because its distribution is discontinued; ENEM Challenge because it is very similar to ENEM 2022, which was already on board; and FaQuAD because its domain is very narrow (viz.~higher education institutions).}

Statistics on the testing datasets are show in Table~\ref{tab:testdatastatistics}.

\begin{table}
    \centering
    \begin{tabular}{lr}
    \toprule
        translated tasks    & \#exs \\ \midrule
        MRPC                & 408           \\
        RTE                 & 277           \\
        COPA                & 100           \\ \midrule
        \multicolumn{1}{r}{subtotal}   & 785           \\
        \bottomrule
        \addlinespace[1ex]
        \toprule
        reused tasks        & \#exs \\ \midrule
        ASSIN2 RTE          & 2448          \\ 
        ASSIN2 STS          & 2448          \\  
        BLUEX               & 178           \\ 
        ENEM 2022           & 118           \\
        FaQuAD              & 63            \\ \midrule
        \multicolumn{1}{r}{subtotal}   & 5256          \\
    \bottomrule
    \end{tabular}
    \caption{Size of translated and reused testing datasets, in number of examples (\#exs). The number of examples is identical for both variants. Reused tasks are pt-br only.}
    \label{tab:testdatastatistics}
\end{table}

\section{Models}
\label{sec:models}

The Gervásio models are based on the LLaMA~2 \citep{Touvron:2023:Llama2} model with 7~billion parameters. 
LLaMA~2 is a open-sourced decoder-based Transformer that has achieved state-of-the-art results in various natural language processing tasks in the English language.
In comparison with previous decoder-based models, such as LLaMA~1 \citep{Touvron:2023:Llama}, the main reasons for its superiority are the use of a larger context length of 4096 tokens and the extensive volume of data it was trained on, a volume that is currently lacking for the Portuguese language.
More specifically, the LLaMA~2 model is pretrained using 2~\emph{trillion} tokens from publicly available sources.
The Gervásio models aim to advance generative AI capacity to handle the Portuguese language by further pretraining it on the data we have curated for Portuguese language variants.

Regarding the details of the decoder architecture, the model has a hidden size of 4096 units, an intermediate size of 11,008 units, 32 attention heads, 32 hidden layers, and a tokenizer obtained using the Byte-Pair Encoding (BPE) algorithm implemented with SentencePiece~\citep{kudo2018sentencepiece}, featuring a vocabulary size of 32,000.

We adopted the LLaMA~2 implementation provided by Hugging Face~\citep{wolf-etal-2020-transformers} as our codebase.
For this purpose, we employed the Transformers library in conjunction with Accelerate~\citep{accelerate}, Flash Attention ~\citep{dao2022flashattention} and DeepSpeed~\citep{rasley2020deepspeed}.

\paragraph{Fine-tuning}

In accordance with the previously described architecture and pre-trained model, we applied supervised fine-tuning for each variant of Portuguese, PTPT and PTBR.
The training objective was causal language modeling (CLM) using the training data specified in Section~\ref{sec:data}.

It is noteworthy that we implemented the zero-out technique during the fine-tuning process, as outlined in \citep{Touvron:2023:Llama2}.
Specifically, while the entire prompt received attention during fine-tuning, only the response tokens were subjected to back-propagation.

In terms of hyper-parameters, we aimed to closely match those utilized in \citep{Touvron:2023:Llama2}.
Consequently, both models were trained with a learning rate of $2 \times 10^{-5}$, a weight decay of 0.1, a two-epoch training regime without warm-up, and to ensure the same number of tokens back-propagated per step, we employed an input sequence of 512 tokens with a batch size of 16 and 16 accumulation steps.

Due to hardware limitations that imposed a shorter sequence length (512) compared to the base model (4096), instead of the typical practice of concatenating all training examples and then dividing them into batches with the same input sequence length, we separated each example individually.
In other words, each example occupies the full input sequence length.

To achieve this, we adapted the tokenizer of the base model to accept padding to allow grouping examples with different size into batches while preserving the original input sequence length.

Considering the substantial discrepancy in dataset sizes between the training set and the pre-training corpus used for the base model, with the latter being orders of magnitude larger, and given the language shift from English to Portuguese, we were uncertain about the expected loss behavior.
We observed that both models exhibited convergence, featuring in the training steps an initial acceleration in terms of loss decay followed by a deceleration.
This behavior suggests the inherent ability of the base model to adapt its focus to a new language, especially considering that the tokenizer was not retrained for Portuguese.

For the model training process, we resorted to an a2-megagpu-16gb Google Cloud A2 VM, equipped with 16 GPUs, 96 vCPUs, and 1.360 GB of RAM. 
The training of each model took approximately two hours.

\section{Evaluation and discussion}
\label{sec:results}

To assess Gervásio models, we resorted to the test sets introduced above in Section~\ref{sec:data}.
For every task under evaluation, we use the respective evaluation metrics commonly found in the literature, typically the F1 score or the Pearson correlation coefficient, as indicated below.

In this connection, it is worth noting that in a text generation task where the generated text is evaluated against a gold label, various responses may arise in which the generated text does not match any of the predefined classes.
In such cases, the response was considered different from the correct label and thus incorrect.
To maintain the integrity of the generated text, which corresponds to the final label, in tasks where the answer is a word, like ``sim'' ou ``não'' (Eng.~``yes'' or ``no''), we only considered the first word provided as the response, after trimming any leading whitespace.
In tasks where the outcome involve classes consisting of single digit numeric value, only the first digit is accepted as the response.

Regarding the hyper-parameters relevant in inference time for the decoder to generate responses to the test tasks, we employed a temperature setting of 1.0, greedy decoding, a beam search value of 1, and applied top-k filtering with a threshold of 50.

Each performance score reported below is the average of the outcome of three independent runs using different seeds.

\begin{table}
    \centering
    \begin{tabular}{lccc}
    \toprule 
        Model           & MRPC      & RTE       & COPA                     \\
        \midrule\addlinespace[1ex]\midrule
        Gervásio ptbr   & \textbf{0.7822}    & \textbf{0.8321}    & 0.2134 \\
        \midrule 
        LLaMA~2         & 0.0369    & 0.0516    & 0.4867   \\
        LLaMA~2 Chat    & 0.5432    & 0.3807    & \textbf{0.5493}   \\
        \midrule\addlinespace[1ex]\midrule
        Gervásio ptpt  & \textbf{0.7273}    & \textbf{0.8291}    & \textbf{0.5459} \\
        \midrule
        LLaMA~2         & 0.0328    & 0.0482    & 0.3844   \\
        LLaMA~2 Chat    & 0.5703    & 0.4697    & 0.4737   \\
    \bottomrule
    \end{tabular}
    \caption{F1 scores for ptbr and ptpt tasks translated from GLUE and SuperGLUE, not seen during training. Best scores for each task are in bold.}
    \label{tab:results}
\end{table}

\paragraph{Tasks from GLUE and SuperGLUE}
\label{sec:gluesupergluetasks}

Each language variant of Gervásio was evaluated with the respective translated version of the test tasks selected from GLUE and SuperGLUE.
The evaluation scores are displayed in Table~\ref{tab:results}.

The LLaMA~2 and LLaMA~2 Chat models were evaluated by us over the Portuguese data for both variants by following also the same approach used for Gervásio, described above.

\begin{table*}
    \centering
    \begin{tabular}{lcccc}
    \toprule
        Model           & ENEM 2022 & BLUEX    & RTE            & STS\\
        \midrule\addlinespace[1ex]\midrule
        Gervásio ptbr   & 0.1977    & 0.2640   & \textbf{0.7469}& \textbf{0.2136}  \\
        \midrule
        LLaMA~2         & 0.2458    & 0.2903   & 0.0913         & 0.1034  \\
        LLaMA~2 Chat    & 0.2232    & 0.2959   & 0.5546         & 0.1750  \\
        \midrule\addlinespace[1ex]\midrule
        Sabiá-7B        & \textbf{0.6017}      & \textbf{0.7743}    & 0.6487         & 0.1363  \\
        %Sabiá-J         & 0.3941    & 0.3989   & 0.6928    & 0.3549         & \textbf{0.2297}  \\
    \bottomrule
    \end{tabular}
    \caption{Evaluation (F1 for RTE, Accuracy for ENEM 2022 and BLUEX, Pearson for STS) in data sets originally developed for American Portuguese, not seen during training. Best scores in bold.}
    \label{tab:results_ptbr}
\end{table*}

\paragraph{Other downstream tasks}
\label{sec:othertasks} 

Gervásio PTBR was also evaluated in the downstream tasks whose data sets were not translated from English but originally developed for Portuguese. The evaluation scores are displayed in Table~\ref{tab:results_ptbr}. For Sabiá, the results presented therte are those reported in the respective publication \citep{pires2023sabia}.
%, obtained under the evaluation used there.

\paragraph{Discussion} \label{sec:discussion} The first important result worth underling is that Gervásio largely outperforms its baseline LLaMA~2 in all tasks by both models, as reported in Table~\ref{tab:results}, except for the PTBR model on the COPA task.

This demonstrates that it was rewarding to continue the causal language modeling of LLaMA~2 with the Portuguese data, even though LLaMA~2 had been pre-trained over a overwhelming majority of English data, and also despite the Portuguese dataset used to continue its pre-training being tiny (1.8~billion tokens) when compared to the one used for LLaMA~2 (2~trillion tokens).\footnote{
To further examine the outlier score of COPA in ptbr, we proceeded with cross evaluation. The PTPT model shown quite similar scores for both the PTPT and PTBR datasets, which seems to indicate that the possible cause for the outlier value did not occur with the construction of the PTBR dataset. The PTBR model, in turn, run over the PTPT testset, shown again an outlier score, similar to the outlier score obtained for PTBR, which may indicate the root of the difference occurs with the training sets. In fact, the base model LLaMA was trained on 2.8 Billion tokens of Portuguese (0.09\% of the total 2 Trillion tokens used for its training in English), where PTBR texts were most probably in much superior number than PTPT ones, given the respective demographics. This indicates in which measure the two training conditions for PTBR and PTPT may differ. Nevertheless, if this larger exposure to PTBR data, by the starting model LLaMA, was the cause for the outlier value with COPA, then it will remain to expalin why the score for MRPC and RTE are in line for both PTBR and PTPT. We leave this for future research.}

Another result from the values in Table~\ref{tab:results} is aligned with similar results that had been found in \citep{Rodrigues:2023:Albertina}.
The different performance scores of Gervásio for each of the language variants reinforce that it is relevant to have a specific version of the model for each language variant.

Turning to Table~\ref{tab:results_ptbr}, one finds the results obtained with datasets originally developed for PTBR, thus not having been obtained by machine translation.
For two of the tasks, namely RTE and STS, the performance scores obtained here repeat the same contrast obtained with the other test datasets translated into Portuguese whereby Gervásio PTBR greatly outperforms its baseline LLaMA~2. 

For the two other tasks, ENEM~2022 and BLUEX, in turn, Gervásio does not show clear advantage over its starting model.
This difference in performance seems to be justified by the different type of tasks in each group.
Gervásio seems to cope better with tasks concerned with comparing sentences (RTE, with binary decision, and STS, with 6-way decision), rather than with tasks concerned with question answering (ENEM2022, with 5-way, and BLUEX, with 4-way), likely less exercised in the training set.

The scores of Sabiá in Table~\ref{tab:results_ptbr} invite to contrast them with Gervásio's but such comparison needs to be taken with some caution.

First, these are a repetition of the scores presented in the respective paper \cite{pires2023sabia}, which only provide results for a single run of each task, while scores of Gervásio are the average of three runs, with different seeds. 

Second, the evaluation methods adopted by Sabiá are \textit{sui generis}, and different from the one's adopted for Gervásio. 
Following Gervásio's decoder nature as a generative model, our scores are obtained by matching the output generated by Gervásio against the ground labels.
Sabiá, in turn, followed a convoluted approach away from its intrinsic generative nature, by ``calculating the likelihood of each candidate answer string based on the input text and subsequently selecting the class with the highest probability'' \cite[p.231]{pires2023sabia}, which forces the answer to be one of the possible classes and likely facilitates higher performance scores than Gervásio's, whose answers are generated without constraints.

Third, to evaluate Sabiá, the examples included in the few-shot prompt are hand picked, and identical for every test instance in each task \cite[p.4]{pires2023sabia}.
To evaluate Gervásio, the examples were randomly selected to be included in the prompts.

%; its results were obtained with a larger number of examples in each few-shot prompt (16), compared to what we used (less than 16, with exact number depending on the length of examples, taking into account our sequence length) \todo{confirmar}.

%Third, Sabiá models were reported to have been trained on a much larger volume of data than the Gervásio models.

%They were pre-trained on approximately 7~billion tokens, whereas the Gervásio models on approximately 83~million tokens. 
%It is worth highlighting that, although they were pretrained using the same training objective as the Gervásio models, Sabiá models did not undergo fine-tuning with instruct data, like Gervásio did, but rather CLM with plain text taken from ``webpages, code, books, and scientific papers'' \cite[\S 3.2]{pires2023sabia}.

Even taking these considerations into account, it is noticeable that the results in Table~\ref{tab:results_ptbr} indicate performance scores for Gervásio that are clearly better than for Sabiá, over the same two test tasks where it also excels over its starting model.

Given that Gervásio, in addition, is distributed as an fully open model, and Sabiá is publicly available for research only, all these circumstances seems to speak for Gervásio's advantage in terms of its usage for research and commercial purposes.

%\todo[inline]{Avaliação cruzada do COPA.
%\\gervásio br avaliado em copa pt:\\seed 41 'f1': 0.24390243902439027\\seed 42 'f1': 0.2077922077922078\\seed 43 'f1': 0.275
%\\gervásio pt avaliado em copa br:\\seed 41 'f1': 0.4705882352941176\\seed 42 'f1': 0.5544554455445544\\seed 43 'f1': 0.49504950495049505}

%\todo[inline]{Limitations: vale a pena dizer que os modelos estão severamente limitados às labels que viram durante o instruct fine-tuning? Podemos oferecer como future work o treino de modelos onde esta limitação pode ser abordada com técnicas como congelar certas layers durante o fine-tuning.\\JS: Acho que vale a pena mencionar isso (e até temos a experiência ``copa12'' para suportar o argumento.}

\section{Conclusion}
\label{sec:conclusion}

This paper contributes new, instruction-tuned large language models of the decoder family of Transformers specifically developed for the Portuguese language, as well as the instruction datasets used to train and evaluate them.

The models are openly available for free and with no registration required under an MIT license at \url{https://huggingface.co/PORTULAN}, where the respective datasets are also openly available for free and with no registration required.

With a 7~billion parameter, these models have an unique set of features for their size.
They are fully open: they are open source; and they are openly distributed, under an open license, thus including for either research or commercial purposes.
They are the most encompassing models for the Portuguese language: they cover both the European variant, spoken in Portugal, and the American variant, spoken in Brazil; and the model for the European variant it is the first of its class, known in the literature.
They show a competitive performance: they outperform other models of similar size publicly reported, thus representing the state of the art. They are fully documented: the new datasets that
were specifically developed for its construction can
be reused and its development can be reproduced; and
reported performance scores can be independently
assessed. 

By being fully open and fully documented, its further development and improvement is openly available to the community.

Also, given their size, these models can still be run on consumer-grade hardware with technological solutions currently available, thus being a contribution to the advancement of research and innovation in language technology for Portuguese.

Future work will include taking these models as the inaugural members of a future family of fully open decoders for Portuguese with a range of other sizes, and characteristics and for other variants of Portuguese.

%\todo[inline]{JS: E as outras variantes de PT? O PT de Angola tem mais falantes, em número absoluto, do que os falantes de PT de Portugal (mas não deve ter dados para treinar modelos). No entanto, se nos posicionamos como um paladino da língua portuguesa, é feio deixar de lado milhões de falantes dessa língua sem sequer lhes fazer menção.}

\section*{Acknowledgements}

This research was partially supported by:
PORTULAN CLARIN---Research Infrastructure for the Science and Technology of Language, funded by Lisboa 2020, Alentejo 2020 and FCT (PINFRA/22117/2016);
ACCELERAT.AI---Multilingual Intelligent Contact Centers, funded by IAPMEI (C625734525-00462629);
and GPTPT---Transformer-based Decoder for the Portuguese Language, funded by FCT (CPCA-IAC/AV/478395/2022).

\section{Annex A: Template example}
\label{sec:templateexample}

As an example, here we describe the template used for the RTE task in PTPT.
In this task, two sentences are given and the task consists in determining whether the first sentence entails the second. 
Each instance in the dataset contains the fields premise, hypothesis and labels.
The template describes how to handle these fields, usually by prepending some string to their contents, as well as defining the initial instruction.

\begin{description}
\item[instruction] ``Nesta tarefa vais receber duas frases. Indica se a primeira frase implica claramente a segunda frase. Ou seja, indica se se conclui que a segunda frase é verdadeira desde que a primeira frase seja verdadeira. Deves responder `sim' se a primeira frase implica a segunda frase ou deves responder `não' no caso contrário.'' (Eng. ``In this task you'll receive two sentences. Indicate whether the first sentence clearly entails the second sentence. That is, indicate whether one can conclude that the second sentence is true as long as the first sentence is true. You should answer `yes' if the first sentence entails the second sentence or `no' otherwise.'')\\
  This is the instruction that is given at the beginning of the input.
\item[premise] ``Frase 1:'' (Eng. ``Sentence 1:'')\\
  This is placed before the contents of the `premise' field of the RTE instance.
\item[hypothesis] ``Frase 2:'' (Eng. ``Sentence 2:'')\\
  This is placed before the contents of the `hypothesis' field of the RTE instance.
\item[pre-label] ``Resposta:'' (Eng. ``Answer:'')\\
  This is placed before the answer.
\item[labels] ``0'' $\to$ ``sim'', ``1'' $\to$ ``não''\\
  This is a mapping from the 0/1 labels used in the RTE dataset to the yes/no labels that are asked for in the instructions for the task.
\end{description}

Applying the template above to an instance gives something like what is shown below.

\begin{center}
\begin{tabular}{p{7.2cm}}
\toprule
Nesta tarefa vais receber duas frases. Indica se a primeira frase implica claramente a segunda frase. Ou seja, indica se se conclui que a segunda frase é verdadeira desde que a primeira frase seja verdadeira. Deves responder `sim' se a primeira frase implica a segunda frase ou deves responder `não' no caso contrário\\
Frase 1: Em 1969, redigiu o relatório que propunha a expulsão do partido do grupo Manifesto. Em 1984, após a morte de Berlinguer, Natta foi eleito secretário do partido.\\
Frase 2: A Natta apoiou o grupo do Manifesto.\\
Resposta: não\\
\bottomrule
\end{tabular}
\end{center}

In addition, a separator string formed by 3 to 5 consecutive `=' (equals) symbols is inserted between each instance in the training data.
And, during few-shot inference, each instance is headed by ``Exemplo $n$'' (Eng. ``Example $n$''), with increasing $n$, and within each instance its few-shot examples are delimited by a separator string formed by 3 or 4 consecutive `-' (hyphen) or `*' (asterisk) symbols.

\section{Annex B: Instruct training tasks}
\label{sec:trainingtasks}

The base tasks and their augmented counterparts that together form the training data are:

\begin{description}
    \item[STS-B] for semantic textual similarity, with augmented \textbf{STS-B Aug1} for generation of a sentence with a STS score of 0/1/2/3/4/5
    \item[WNLI] for coreference and natural language inference, with augmented \textbf{WNLI Aug1} for generating an hypothesis with Positive/Negative inference, and \textbf{WNLI Aug2} for generating a premise with Positive/Negative inference
    \item[BoolQ] for Yes/No question answering, with augmented \textbf{BoolQ Aug1} for question generation with Yes/No answer based on an excerpt, and \textbf{BoolQ Aug2} for excerpt generation with Yes/No answer to a question
    \item[CB] for inference with labels Entailment (E), Contradiction (C) and Neutral (N), with augmented \textbf{CB Aug1} for generating an hypothesis with label E/C/N, and \textbf{CB Aug2} for generating a premise with label E/C/N
    \item[MultiRC] for question answering, with augmented \textbf{MultiRC Aug1} for question generation, \textbf{MultiRC Aug2} for excerpt generation, and \textbf{MultiRC Aug3} for answer generation
\end{description}

\section{Bibliographical References}
\vspace{-3ex}  % hack to remove extra vertical space

\bibliographystyle{lrec-coling2024-natbib}
\bibliography{bibliographic_resources}

\section{Language Resource References}

\citetlanguageresource{*}
\vspace{-8ex} % hack to remove extra vertical space (seemingly introduced by \citetlanguageresource?)

\bibliographystylelanguageresource{lrec-coling2024-natbib}
\bibliographylanguageresource{language_resources}

\end{document}